\documentclass[twoside,11pt]{article}

\usepackage[abbrvbib,preprint]{jmlr2e}

\usepackage{amsmath,amssymb,amsfonts}
\usepackage{mathtools}
\usepackage{algorithmic}
\usepackage{algorithm}
\usepackage{subcaption}
\usepackage{graphicx}
\usepackage{textcomp}
\usepackage{xcolor}
\usepackage[]{listofsymbols}
\usepackage{siunitx}
\usepackage{tikz}
\usepackage[utf8]{inputenc}
\usepackage{pgf}
\usepackage{pgfplots} 
\usepackage{pgfgantt}
\usepackage{pdflscape}
\pgfplotsset{compat=1.18}
\pgfplotsset{plot coordinates/math parser=false}
\usepackage{bm}
\usetikzlibrary{arrows, positioning, calc, arrows.meta, shapes.multipart, patterns, bending}
\usetikzlibrary{plotmarks}
\usepgfplotslibrary{patchplots}
\usepgfplotslibrary{fillbetween}
\usepgfplotslibrary{groupplots}
\usepackage{grffile}
\usepackage{etoolbox}
\usepackage{ifthen}

\usepackage{tabularx,booktabs}
\usepackage{multirow}

\definecolor{blau}{rgb}{0.1216,0.44700,0.7059}%
\definecolor{rot}{rgb}{0.85000,0.32500,0.09800}%
\definecolor{orange}{rgb}{1.0,0.4980,0.0549}%
\definecolor{lila}{rgb}{0.4940,0.1840,0.5560}%
\definecolor{gruen}{rgb}{0.1725,0.62745,0.1725}%
\definecolor{hellblau}{rgb}{0.3010,0.7450,0.9330}%
\definecolor{dunkelrot}{rgb}{0.6350,0.0780,0.1840}%
\definecolor{hellrot}{rgb}{1,0,0}%
\definecolor{magenta}{rgb}{1,0,1}%

\newcommand{\Normal}[2]{\mathcal{N}\left({#1},{#2}\right)}

\usepackage{lastpage}

\ShortHeadings{Data Reduction Criteria for Gaussian Processes}{Wietzke and Graichen}
\firstpageno{1}

\begin{document}

\title{Comparison of Data Reduction Criteria for Online Gaussian Processes}

\author{\name Thore Wietzke \email Thore.Wietzke@fau.de \\
       \addr Chair of Automatic Control\\
       Friedrich-Alexander University of Erlangen–Nuremberg\\
       Cauerstraße 7, 91058 Erlangen, Germany
       \AND
       \name Knut Graichen \email Knut.Graichen@fau.de \\
       \addr Chair of Automatic Control\\
       Friedrich-Alexander University of Erlangen–Nuremberg\\
       Cauerstraße 7, 91058 Erlangen, Germany}

\editor{My editor}

\maketitle

\begin{abstract}
    Gaussian Processes (GPs) are widely used for regression and system identification due to their flexibility and ability to quantify uncertainty. 
    However, their computational complexity limits their applicability to small datasets.
    Moreover, in a streaming scenario more and more datapoints accumulate which is intractable even for Sparse GPs.
    Online GPs aim to alleviate this problem by e.g. defining a maximum budget of datapoints and removing redundant datapoints.
    This work provides a unified comparison of several reduction criteria, analyzing both their computational complexity and reduction behavior. 
    The criteria are evaluated on benchmark functions and real-world datasets, including dynamic system identification tasks. 
    Additionally, acceptance criteria are proposed to further filter out redundant datapoints.
    This work presents practical guidelines for choosing a suitable criterion for an online GP algorithm.
\end{abstract}


\section{Introduction}

Gaussian Processes (GPs) are a powerful and flexible class of non-parametric models widely used in machine learning for regression, classification, and other tasks like Bayesian optimization~\citep{Rasmussen2005}.
Since their modeling is stochastic in nature, GPs provide uncertainty estimates for the posterior, which makes them especially suitable for stochastic system identification, where many sources of uncertainty like noise or disturbances occur.

Despite their advantages, the computational complexity of GPs scales with $\mathcal{O}(N^3)$, where $N$ is the number of datapoints, which becomes intractable for large datasets.
To alleviate this problem, various sparse approximations have been proposed.
The current state of the art are so-called inducing point methods, where a small set of $M$ points approximates the full GP, resulting in a computational complexity of $\mathcal{O}(MN^2)$.
In this context, the Fully Independent Training Conditional (FITC)~\citep{Snelson2005} and Variational Free Energy (VFE)~\citep{Titsias2009} frameworks are the most commonly used approximations. 

Many real-world regression tasks continuously produce new datapoints, which can be included in the regression model.
Such online settings with streaming data quickly get intractable if every datapoint shall be included.
Therefore, most online approaches impose a maximum number or budget of datapoints.
When this budget is exceeded, different strategies are used to manage the inclusion of new datapoints.
The first class of online GPs are projection-based algorithms.
A popular approach from~\citep{Csato2002} employs a set of basis vectors, similar to the inducing inputs in sparse GPs.
If a new input is within an insertion tolerance, it is projected onto the existing dataset; otherwise, it replaces an existing basis vector.
A related method~\citep{Le2017}, which does not remove an inducing input, provides a geometric interpretation.
Both approaches use the predictive variance as a criterion for insertion and reduction.
An alternative~\citep{Bijl2015} uses the FITC approximation to project new datapoints directly onto the inducing mean and variance of the GP, independent of any insertion criterion.
Other methods utilize Kalman filter-like updates for GPs~\citep{Huber2014} or sparse GPs~\citep{Schuerch2020}.

Another class of methods involves direct insertion and reduction without projection.
For example,~\citep{Nguyen2011} uses the predictive variance to prune redundant datapoints and derives explicit update rules for the GP covariance matrix, including a temporal weighting scheme for reduction.
This approach has also been combined with the FITC approximation in~\citep{Kabzan2019}, where it was experimentally validated on an autonomous racecar.
The data selection in~\citep{Kabzan2019} was performed in a parallel process from the main control process to ensure real-time capabilities.
Another approach in~\citep{Petelin2013} uses the log marginal likelihood of the GP as a reduction criterion.
These reduction strategies can also be used to choose the most influential datapoints for the GP.
One work in that direction is~\citep{Bergmann2018}, which reduced the number of datapoints in removing the datapoints with the least impact on the mean prediction.

Another related field to datapoint reduction is active learning.
The goal is inverse to reduction, as it is concerned about which datapoint to include.
Common approaches use the predictive entropy or mutual information to choose the datapoints~\citep{Krause2008}.
For system identification this yields an optimal control problem where e.g. the entropy is maximized over a horizon~\citep{Fenet2020}.

The discussion shows that various strategies in different areas have been proposed for selecting the number of datapoints.
But no comparison of the current approaches was conducted so far.
This work aims to close this gap and looks at the different criterions from a unified angle.
Our contributions are:
\begin{itemize}
    \item We review and compare the data reduction approaches. 
    \item We present alternative formulations which either provide a computational advantage or provide a better understanding of the approaches.
    \item Since insertion and reduction criteria are different, we added an acceptance criterion for every reduction approach to filter out unsuitable datapoints.
    \item We provide practical guidelines for the usage of the reduction and acceptance criteria.
\end{itemize}

The paper is outlined as follows:
First, we provide a background to Gaussian Process Regression in Section~\ref{sec:GaussianProcess}.
Then, the online GPs are introduced in Section~\ref{sec:OnlineGaussianProcess}, including the insertion, reducing and our proposed acceptance criteria.
In Section~\ref{sec:OfflineEvaluation} the reduction and acceptance criterions are evaluated and compared in an offline setting.
Finally, Section~\ref{sec:OnlineEvaluation} evaluates the reduction and acceptance criteria within an online GP with two benchmarks.

\section{Gaussian Processes for Regression}
\label{sec:GaussianProcess}

Gaussian Process Regression is a versatile non-parametric approach for modeling an unknown function $f(\bm x)$~\citep{Rasmussen2005}.
The training dataset is represented as $\mathcal{D} = (\bm{X}, \bm{y})$, where $\bm{X} \in \mathbb{R}^{n \times p}$ is a matrix containing $n$ input feature vectors $\bm{x}_i \in \mathbb{R}^p$, and $\bm{y} \in \mathbb{R}^n$ is a vector of corresponding target values.
The targets are typically modeled as noisy observations of the function, expressed as
\begin{equation}
    \label{eq:M_y=f}
    y = f(\bm x) + \epsilon, \quad \epsilon \sim \Normal{0}{\sigma_\epsilon^2},
\end{equation}%
where $\epsilon$ represents Gaussian noise with zero mean and variance $\sigma_\epsilon^2$.
The zero-mean GP prior for the latent function values $\bm{f} = f(\bm{X})$ is given by
\begin{equation}
    p(\bm{f}) = \Normal{\bm{0}}{\bm{K}_{ff}},
\end{equation}%
where $\bm{K}_{ff}$ is the covariance matrix computed using the kernel function $k(\bm{x}, \bm{x}')$.
The noisy observations $\bm{y}$ are modeled as a Gaussian distribution conditioned on the latent function values with 
\begin{equation}
    p(\bm{y}|\bm{f}) = \Normal{\bm{0}}{\bm{K}_{ff}+\sigma_\epsilon^2\bm{I}}.
\end{equation}%
To predict the function values $f(\bm{x}_*)$ at test inputs $\bm{X}_*$, we assume a joint Gaussian distribution $p(\bm{y},\bm{f}_*)$ with
\begin{equation}
    \label{eq:JointDistribution}
    \begin{bmatrix}
        \bm{y} \\ \bm{f}_*
    \end{bmatrix}
    \sim \Normal{
    \begin{bmatrix}
        \bm{0} \\ \bm{0}
    \end{bmatrix}}{
    \begin{bmatrix}
        \bm{K}_{ff}+\sigma_\epsilon^2\bm{I} & \bm{K}_{f*} \\
        \bm{K}_{*f} & \bm{K}_{**}
    \end{bmatrix}},
\end{equation}%
where $\bm{K}_{f*}$ and $\bm{K}_{*f}$ are the cross-covariance matrices, and $\bm{K}_{**}$ is the covariance matrix for the test inputs. 
The predictive distribution $p(\bm{f}_*|\bm{y})$ is Gaussian with mean and covariance given by
\begin{align}
    \bm{\mu}_* &= \bm{K}_{*f} \left(\bm{K}_{ff}+\sigma_\epsilon^2\bm{I}\right)^{-1} \bm{y}, \label{eq:pred_mu}\\
    \bm{\Sigma}_* &= \bm{K}_{**} - \bm{K}_{*f} \left(\bm{K}_{ff}+\sigma_\epsilon^2\bm{I}\right)^{-1} \bm{K}_{f*}. \label{eq:pred_Sigma}
\end{align}

The kernel function and the GP are dependent on some hyperparameters $\bm{\theta}$. 
They can be computed from the maximum log likelihood of $p(\bm y)$ given by
\begin{align}
    \log(p(\bm y)) &= - \frac{1}{2}\bm{y}^T\left(\bm{K}_{ff} + \sigma_\epsilon^2\bm{I}\right)^{-1}\bm{y} - \frac{1}{2}\log\left|\bm{K}_{ff} + \sigma_\epsilon^2\bm{I}\right| - \frac{n}{2}\log(2\pi)
\end{align}
where $\left|.\right|$ denotes the determinant of a matrix.
The maximization is typically done by gradient descend~\citep{Bishop2006}.

A commonly used kernel function is the squared exponential (SE) given by
\begin{equation}
    k(\bm{x}, \bm{x}') = \sigma_s \exp\left(-\frac{1}{2}(\bm{x}-\bm{x}')^T\bm{\Lambda}^{-1}(\bm{x}-\bm{x}')\right)
\end{equation}
where $\sigma_s$ is the signal variance and $\bm{\Lambda} = \mathrm{diag}(\bm{l}^2)$ with the lengthscales $\bm{l}$.

Gaussian Processes can also be understood from a geometric perspective, particularly through their connection to kernel methods and the theory of Reproducing Kernel Hilbert Spaces (RKHS). 
Similar to Support Vector Machines (SVMs), GPs rely on a mapping $\Phi(\bm x)$ into a (possibly infinite-dimensional) Hilbert space $\mathcal{H}$, specifically the RKHS, associated with a positive-definite kernel function $k(\bm x, \bm x')$. 
In this space, inner products between mapped inputs can be expressed as
\begin{equation}
    k(\bm x, \bm x') = \langle\Phi(\bm x), \Phi(\bm x')\rangle_\mathcal{H},
\end{equation}
a principle known as the \emph{kernel trick}, which allows computations in high-dimensional spaces to be carried out implicitly.

From a regression perspective, the feature map $\Phi(\bm x)$ can be seen as defining a (potentially infinite) set of basis functions. 
By specifying a GP with a kernel that induces an infinite-dimensional RKHS, the model implicitly performs Bayesian linear regression in this rich function space. 
For a detailed introduction to the RKHS and its role in machine learning, see~\citep{Scholkopf2002}.





\section{Online Gaussian Processes}
\label{sec:OnlineGaussianProcess}

As pointed out in the introduction, using a GP for a streaming or online regression task is intractable, since at one point the maximum available memory is exceeded.
Thus, a maximum budget of datapoints $N_{max}$ is defined.
To account for the maximum datapoint budget, insertion and reduction criteria are needed to first select suitable datapoints and the remove an existing datapoint.

The basic structure of the online GP algorithm is outlined in Algorithm~\ref{alg:online} and is heavily inspired by~\citep{Petelin2013}.
One starts with an initialized GP with the dataset $\mathcal{D}$ of size $N = \left|\mathcal{D}\right|$, where hyperparameter training was already conducted.
If a new measurement $\mathcal{D}_*$ arrives and if the new datapoint meets a suitable insertion criterion, the datapoint is added if the budget $N_{max}$ is not exceeded.
Otherwise, an existing datapoint has to be replaced, for which an acceptance criterion is checked.
If accepted, the reduction determines the datapoint $\mathcal{D}_r$ of $\mathcal{D}$ which shall be replaced with $\mathcal{D}_*$, which is denoted by $\mathcal{D} = \left\{\mathcal{D}\:|\:\mathcal{D}_r = \mathcal{D}_*\right\}$.
Afterwards, if $\mathcal{D}$ was updated, precomputable quantities of the GP are cached for computing the predictive density in~\eqref{eq:pred_mu} and~\eqref{eq:pred_Sigma}.

\begin{algorithm}
    \caption{Online Gaussian Process}
    \label{alg:online}
    \begin{algorithmic}
        \STATE $\mathcal{D} = (\bm{X}, \bm{y})$
        \STATE $N = \left|\mathcal{D}\right|$
        \STATE set $N_{max}$
        \LOOP
            \STATE Measure new datapoint $\mathcal{D}_* = (\bm x_*, y_*)$
            \IF{\texttt{insert\_datapoint($\mathcal{D}_*$)}}
                \IF{$N < N_{max}$}
                    \STATE $\mathcal{D} \leftarrow \mathcal{D} \cup \mathcal{D}_*$
                    \STATE $N = \left|\mathcal{D}\right|$ 
                    \STATE \texttt{cache\_GP()}
                \ELSE
                    \IF{\texttt{accept\_datapoint($\mathcal{D}_*$)}}
                        \STATE $r = \texttt{reduce\_datapoint($\mathcal{D}_*$)}$
                        \STATE $\mathcal{D} = \left\{\mathcal{D}\:|\:\mathcal{D}_r = \mathcal{D}_*\right\}$
                    \ENDIF
                \ENDIF
            \ENDIF
        \ENDLOOP
    \end{algorithmic}
\end{algorithm}

In the following sections, the insertion, reduction and acceptance criteria are discussed.
The acceptance criteria are introduced following the reduction criteria, since they are conceptually derived from them.

\subsection{Insertion criteria}

For an online GP, insertion criteria form the first filter to find suitable datapoints.
The first occurrence of such criterion for GPs can be found in~\citep{Csato2002} with further refinement in~\citep{Nguyen2011}.
The common ground is to compute the distance of a new point $\bm{x}_*$ to $\mathcal{D}$.
Using the RKHS, the so-called rejection vector is given by
\begin{equation}
    \bm r = \Phi(\bm x_*) - \sum_{i=1}^n \alpha_i \Phi(\bm x_i),
\end{equation}
where $\alpha_i$ are the entries of $\bm \alpha = \left(\bm{K}_{ff} + \sigma_\epsilon^2\bm{I}\right)^{-1}\bm{K}_{f}$.
The Euclidean norm of $\bm r$ corresponds to the predictive standard deviation $\sigma_*$ of the GP and is computed by
\begin{align}
    \sigma_* = \left\|\bm r\right\| &= \big|\big| \Phi(\bm x_*) - \sum_i \alpha_i \Phi(\bm x_i)\big|\big| \\
    &= \sqrt{K_{**} - \bm{K}_{*f} \left(\bm{K}_{ff}+\sigma_\epsilon^2\bm{I}\right)^{-1} \bm{K}_{f*}}.
\end{align}
A new datapoint will be included if $\sigma_*^2 > \bar{\sigma}^2$ for some variance threshold $\bar{\sigma}^2 > 0$.
Another reasonable insertion criterion is the absolute prediction error $\left|y - \mu(\bm x_*)\right| > \bar{e}$ for an error threshold $\bar{e}$~\citep{Petelin2013}. 
The combination of both criteria as considered in~\citep{Maiworm2018} is outlined in Algorithm~\ref{alg:insertion}.

\begin{algorithm}
    \caption{\texttt{insert\_datapoint()}}
    \label{alg:insertion}
    \begin{algorithmic}
        \STATE $\mathcal{D}_* = (\bm x_*, y_*)$
        \IF{$\sigma_*^2 > \bar{\sigma}^2$ \OR $\left|y_* - \mu_*\right| \geq \bar{e}$} 
            \RETURN \TRUE
        \ELSE
            \RETURN \FALSE
        \ENDIF
    \end{algorithmic}
\end{algorithm}

\subsection{Reduction Criteria}

After the insertion criterion found a suitable datapoint $\mathcal{D}_*$, the GP has to remove an existing datapoint if the maximum budget $N_{max}$ is reached.
This step requires to evaluate a suitable acceptance and reduction criterion to quantify which point to remove.
As mentioned before, the acceptance criteria are influenced by the reduction criteria, thus they are introduced afterwards.

The fundamental algorithm is outlined in Algorithm~\ref{alg:reduction}. 
Each datapoint $\mathcal{D}_i$ of the set $\mathcal{D}$ is sequentially replaced with $\mathcal{D}_*$, and $f_{\mathrm{red}}(\mathcal{D}_{\!/i})$ is evaluated, where the index $/i$ indicates the dataset with $\mathcal{D}_*$, but without $\mathcal{D}_i$.
After evaluating all datapoints, the datapoint with the index $r$ corresponding to the lowest score is replaced with $\mathcal{D}_*$.

\begin{algorithm}
    \caption{\texttt{reduce\_datapoint()}}
    \label{alg:reduction}
    \begin{algorithmic}
        \STATE $N = \left|\mathcal{D}\right|$
        \STATE $J \in \mathbb{R}^N$
        \STATE $\mathcal{D}_* = (\bm x_*, y_*)$
        \FOR{$i$ \TO $N$}
            \STATE $\mathcal{D}_{\!/i} = \left\{\mathcal{D}\:|\:\mathcal{D}_i = \mathcal{D}_*\right\}$
            \STATE $J(i) = f_{\mathrm{red}}(\mathcal{D}_{\!/i})$
        \ENDFOR
        \STATE $r = \mathrm{arg}\mathrm{min} \: J(i)$
        \RETURN $r$
    \end{algorithmic}
\end{algorithm}

In the following, suitable reduction criteria are proposed.
They are based on information theoretic metrics and heuristics.

\subsubsection{Predictive Entropy}

The first criterion for data reduction is the predictive entropy of the GP,  which is closely related to the predictive variance.
The predictive variance was already used for data reduction in~\citep{Nguyen2011,Kabzan2019}.
Entropy was originally introduced by Shannon for discrete random variables to quantify the average information content. 
For continuous random variables, the analogous concept is differential entropy~\citep{Mackay2003}, which, unlike discrete entropy, can take negative values. 
In this context, we refer to differential entropy simply as entropy for brevity, since only continuous random variables are considered. 

For a probability density function $p(x)$ of a continuous random variable $x$, the entropy is defined as
\begin{equation}
    H(p(x)) = \mathbb{E}\left[-\log(p(x))\right] = -\int_{-\infty}^{\infty} p(x)\log(p(x))\mathrm{dx}.
\end{equation}%
For a multivariate normal distribution $p(\bm x) = \Normal{\bm \mu}{\bm \Sigma}$, the entropy is computed by
\begin{equation}
    H(p(\bm x)) = \frac{n}{2}(1+\log(2\pi)) + \frac{1}{2}\log\left|\bm \Sigma\right|.
\end{equation}%
Inserting the predictive distribution $p(\bm{f}_*|\bm{y})$, which is also Gaussian, yields the \emph{Predictive Entropy}
\begin{equation}
    \label{eq:pred_Entropy}
    f_{\mathrm{red}}(\mathcal{D}_{\!/i}) = H(p(f_i|\bm y_{/i})) = \frac{1}{2}(1+\log(2\pi)) + \frac{1}{2}\log(\sigma_i^2),
\end{equation}%
where $\sigma_i^2$ represents the predictive variance for the $i$th test point $\bm x_i$. 
A smaller predictive variance and consequently lower entropy indicates higher confidence in the predicted value, suggesting that the prediction is well-supported by the existing datapoints.
Contrary to active learning~\citep{Krause2008}, a small entropy is desired since $\bm x_*$ is already included.
From a geometric viewpoint, computing the distance of $\Phi(\bm x_i)$ to the span $\Phi(\bm X)$ is the predictive standard deviation $\sigma_i$.
Thus, a smaller \emph{Predictive Entropy} yields a smaller distance to the spanned hyperplane in the RKHS.

A more efficient way to compute the \emph{Predictive Entropy}~\eqref{eq:pred_Entropy} can be retrieved when considering the joint distribution $p(\bm{y},\bm{f}_*)$ given in~\eqref{eq:JointDistribution}. 
The determinant of the joint covariance matrix can be expressed as
\begin{align}
    \label{eq:JointDeterminant}
    \left|\bm\Sigma\right| &= \left|\bm{K}_{ff} + \sigma_\epsilon^2\bm{I}\right|\left|\bm{K}_{**} - \bm{K}_{*f} \left(\bm{K}_{ff}+\sigma_\epsilon^2\bm{I}\right)^{-1} \bm{K}_{f*}\right| \nonumber \\
    &= \left|\bm{K}_{ff} + \sigma_\epsilon^2\bm{I}\right|\left|\bm{\Sigma}_*\right| 
\end{align}
using the block matrix determinant identity~\citep{MatrixCookbook}.
The second term corresponds to the predictive variance $\Sigma_*$ given in~\eqref{eq:pred_Sigma}, which is used in the \emph{Predictive Entropy}.
If the predictive variance $\bm{\Sigma}_*$ is lower for a particular partition, the determinant of the prior covariance $\bm{K}_{ff_{/i}}$ becomes higher, since the joint determinant $\left|\bm\Sigma\right|$ remains constant.
Thus, minimizing the predictive variance is equivalent to maximizing the determinant of the prior covariance $\bm{K}_{ff_{/i}}$.
This yields the \emph{Prior Entropy} criterion given by
\begin{equation}
    \label{eq:prior_Entropy}
    f_{\mathrm{red}}(\mathcal{D}_{\!/i}) = -H(p(\bm y_{/i})),
\end{equation}
with
\begin{equation}
    H(p(\bm y_{/i})) = \frac{n}{2}(1+\log(2\pi)) + \frac{1}{2}\log\left|\bm{K}_{ff_{/i}}\right|
\end{equation}
as a computational efficient alternative to~\eqref{eq:pred_Entropy}, since $\sigma_i^2$ does not need to be computed.

At first glance, maximizing the entropy may seem counterintuitive as lower entropy implies lower uncertainty.
However, a lower \emph{Prior Entropy} indicates a more concentrated distribution, meaning the datapoints are highly similar, and the explored region is small.
Conversely, a higher entropy corresponds to a broader explored region, which is generally preferable for a GP prior.
Geometrically, this relates to the volume of the dataset in the RKHS given by
\begin{equation}
    \mathrm{vol}(\Phi(\bm X)) = \sqrt{\det(\langle\Phi(\bm X),\Phi(\bm X)\rangle_\mathcal{H})} = \sqrt{\left|\bm K_{ff_{/i}}\right|}.
\end{equation}

Since the \emph{Predictive Entropy}~\eqref{eq:pred_Entropy} and \emph{Prior Entropy}~\eqref{eq:prior_Entropy} are coupled via~\eqref{eq:JointDeterminant}, minimizing~\eqref{eq:pred_Entropy} results in the maximization of~\eqref{eq:prior_Entropy} and vice versa.
Thus, \emph{Predictive Entropy} and \emph{Prior Entropy} can be seen as the same criterion, while~\eqref{eq:prior_Entropy} is faster to compute.

\subsubsection{Mean Relevance}

A heuristic approach to reduce the number of datapoints is the \emph{Mean Relevance}~\citep{Bergmann2018}
\begin{equation}
    \label{eq:MR}
    f_{\mathrm{red}}(\mathcal{D}_{\!/i}) = (\mu_i - \mu_{/i})^2
\end{equation}
that measures the impact on the predictive mean~\eqref{eq:pred_mu} for $\bm{x}_i$ when removing the ith datapoint. 
Thus, a lower score indicates less impact for $\mu_i$ and thus a lower relevance.

\subsubsection{Marginal Log Likelihood}

In the work of~\citep{Petelin2013}, the information gain from including the $i$th datapoint is evaluated using the \emph{Marginal Log Likelihood} of the GP, defined as
\begin{multline}
    \label{eq:MLL}
    f_{\mathrm{red}}(\mathcal{D}_{\!/i}) = \log(p(\bm y_{/i})) = - \frac{1}{2}\bm{y}_{/i}^T\left(\bm{K}_{ff_{/i}} + \sigma_\epsilon^2\bm{I}\right)^{-1}\bm{y}_{/i}
    - \frac{1}{2}\log\left|\bm{K}_{ff_{/i}} + \sigma_\epsilon^2\bm{I}\right| - \frac{n}{2}\log(2\pi).
\end{multline}
A lower \emph{Marginal Log Likelihood} indicates that the excluded datapoint contains less information.
This can be clarified by considering the joint distribution $p(\bm{y}) = p(\bm{y}_{/i},y_i) = p(y_i|\bm{y}_{/i})p(\bm{y}_{/i})$.
Taking the logarithm yields 
\begin{equation}
    \log(p(\bm y)) = \log(p(\bm y_{/i})) + \log(p(y_i|\bm y_{/i}))
\end{equation}
with 
\begin{align}
    \log(p(y_i|\bm y_{/i})) &= -\frac{1}{2}\log(2\pi(\sigma_i^2+\sigma_\epsilon^2)) - \frac{(y_i - \mu_i)^2}{2(\sigma_i^2+\sigma_\epsilon^2)},
\end{align}%
where $\sigma_i^2$ is the variance at point $\bm x_i$. 
The \emph{Log Predictive Density} criterion is given by
\begin{equation}
    \label{eq:LPD}
    f_{\mathrm{red}}(\mathcal{D}_{\!/i}) = -\log(p(y_i|\bm y_{/i})).
\end{equation}
For the \emph{Log Predictive Density}, a higher value indicates a better stochastic fit, as lower variance and smaller prediction error $(y_i - \mu_i)^2$ both increase the \emph{Log Predictive Density}.

Since the joint distribution $p(\bm{y})$ is constant, a lower \emph{Marginal Log Likelihood} for a given $i$ results in a higher \emph{Log Predictive Density}.
Thus,~\eqref{eq:MLL} and~\eqref{eq:LPD} are equivalent in their results.
The \emph{Marginal Log Likelihood} is faster to compute since the predictive mean and variance in~\eqref{eq:pred_mu} and~\eqref{eq:pred_Sigma} do not have to be computed.
Note that \emph{Marginal Log Likelihood} and \emph{Log Predictive Density} can be interpreted as combining aspects of the \emph{Mean Relevance}~\eqref{eq:MR} and \emph{Predictive Entropy}~\eqref{eq:pred_Entropy}.

\subsection{Acceptance Criteria}
\label{sec:AcceptanceCriteria}

When different criteria are used for insertion and reduction, the insertion criterion may propose datapoints that are removed by the reduction criterion in the next step. 
This can lead to repeated reevaluation and removal of recently added points in Algorithm~\ref{alg:online}. 
To address this, we introduce an acceptance criterion $f_{\mathrm{acc}(\mathcal{D}_*)}$ corresponding to each reduction criterion, to filter proposed datapoints by Algorithm~\ref{alg:insertion}.

The algorithm is shown in Algorithm~\ref{alg:acceptance}.
First, the acceptance criterion $f_{\mathrm{acc}}(.)$ is evaluated for every datapoint $\mathcal{D}_i$ in $\mathcal{D}$, where $i$ denotes the datapoint index.
Afterwards, the minimum of $J$ is computed.
If $f_{\mathrm{acc}}(\mathcal{D}_*)$ exceeds $J_{min}$, the new datapoint is accepted and the replace algorithm is executed.
Note that the computation of $J_{min}$ is formally included in \texttt{accept\_datapoint()}, but should be computed during \texttt{cache\_GP()} to avoid unnecessary reevaluations.

\begin{algorithm}
    \caption{\texttt{accept\_datapoint()}}
    \label{alg:acceptance}
    \begin{algorithmic}
        \STATE $N = \left|\mathcal{D}\right|$
        \STATE $\bm{J} = \left[J_1, ..., J_N\right]^T$
        \STATE $\mathcal{D}_* = (\bm x_*, y_*)$
        \FOR{$i$ \TO $N$}
            \STATE $J_i = f_{\mathrm{acc}}(\mathcal{D}_i)$
        \ENDFOR
        \STATE $J_{\mathrm{min}} = \mathrm{min} \: \bm{J}$
        \IF{$f_{\mathrm{acc}}(\mathcal{D}_*) > J_{\mathrm{min}}$} 
            \RETURN \TRUE
        \ELSE
            \RETURN \FALSE
        \ENDIF
    \end{algorithmic}
\end{algorithm}

\subsubsection{Predictive Entropy}

For the \emph{Predictive Entropy}~\eqref{eq:pred_Entropy}, the acceptance criterion is given by the predictive variance in~\eqref{eq:pred_Sigma}.
If~\eqref{eq:pred_Sigma} for a given point is lower than the minimum predictive variance of $\mathcal{D}$, then this point is already well covered by the current dataset and the gained information content is low.
The resulting acceptance criterion can be computed by
\begin{equation}
    f_{\mathrm{acc}}(\mathcal{D}_i) = \sigma^2(\bm{x}_i).
\end{equation}

\subsubsection{Mean Relevance}

For the \emph{Mean Relevance}~\eqref{eq:MR} we look at the prediction error given by
\begin{equation}
    f_{\mathrm{acc}}(\mathcal{D}_i) = (y_i - \mu_i)^2
\end{equation}
where $\mu_i$ is given by~\eqref{eq:pred_mu}.
If the prediction error for the new datapoint is higher than the minimum prediction error among the existing datapoints, the new point is accepted.

\subsubsection{Marginal Log Likelihood}

For the \emph{Marginal Log Likelihood}~\eqref{eq:MLL}, we utilize the similarity to the \emph{Log Predictive Density}~\eqref{eq:LPD}.
The criterion is given by
\begin{align}
    f_{\mathrm{acc}}(\mathcal{D}_i) = -\log(p(y_i|\bm y_{/i})) &= \frac{1}{2}\log(2\pi(\sigma_i^2+\sigma_\epsilon^2)) + \frac{(y_i - \mu_i)^2}{2(\sigma_i^2+\sigma_\epsilon^2)}
\end{align}
which is the negative \emph{Log Predictive Density}. 
This acceptance criterion is a combination of the \emph{Predictive Entropy}~\eqref{eq:pred_Entropy} and the \emph{Mean Relevance}~\eqref{eq:MR}, as both the prediction error and the variance is considered.
A new datapoint can be accepted, if either the variance or the prediction error is higher than the current minimum of the negative \emph{Log Predictive Density}.

\section{Benchmark Functions and Datasets}
\label{sec:Benchmarks}
For our evaluation we use several benchmark functions and datasets for standard regression tasks as well as for system identification tasks.
For a better readability, every dataset name is typed in boldface.
The code to generate the data for the benchmark functions and the new system identification tasks is available at \url{https://github.com/ThoreWietzke/Data-Reduction-Criteria-Gaussian-Process-Datasets}.

\subsection{Benchmark Functions}
The first category are test functions of four well-known optimization benchmarks.
The \textbf{Himmelblau} function~\citep{Himmelblau1972} is defined as 
\begin{equation}
    f(x_1, x_2) = (x_1^2 + x_2 - 11)^2 + (x_1 + x_2^2 - 7)^2
\end{equation}
and is evaluated on a grid with $x_1, x_2 \in [-4, 4]$. 
The \textbf{Rastrigin} function~\citep{yang2010} given by 
\begin{equation}
    f(x_1, x_2) = 20 + (x_1^2 - 10\cos(2\pi x_1)) + (x_2^2 - 10\cos(2\pi x_2))
\end{equation}
is evaluated for $x_1, x_2 \in [-1, 1]$. 
The \textbf{Six Hump Camel} function~\citep{yang2010} is defined with
\begin{equation}
    f(x_1, x_2) = (4-2.1 x_1^2 + \frac{x_1^4}{3})x_1^2 + x_1 x_2 + (-4+4x_2)x_2^2
\end{equation}
on a grid with $x_1, x_2 \in [-2, 2]$. 
Finally, the \textbf{Rosenbrock} function~\citep{yang2010}
\begin{equation}
    f(x_1, x_2) = (1-x_1)^2 + 100(x_2-x_1^2)^2
\end{equation}
is also evaluated for $x_1, x_2 \in [-1, 1]$.
These benchmark functions provide easy visualization, validation and sampling for the evaluation.

\subsection{Static Benchmark Datasets}
Two publicly available benchmark datasets are used to compare the reduction criteria.
The first one is the well-known \textbf{Boston} Housing set~\citep{BostonHousing}. 
It consists of 506 datapoints with 13 input features and one target value, either the nitrous oxide level or the price of the houses. 

The other dataset is the \textbf{Concrete} Compressive Strength set~\citep{Concrete} with 8 input features, 1030 instances and one target value, the compressive strength.
For both datasets the input features were normalized and the mean of the target value was removed.

\subsection{System Identification Benchmark Datasets}
The remaining four datasets are dynamic systems based on differential equations.
The regression task for the dynamic systems is given by
\begin{equation}
    \label{eq:DynamicTarget}
    y_k = y_{k-1} + f(y_{k-1}, ..., y_{k-n_y}, u_{k-1}, ..., u_{k-n_u}),
\end{equation}
where $f$ corresponds to the change in the output, $\Delta y_k = y_k - y_{k-1}$.
The number of lags for $n_y$ corresponds to the system order.
The lags for $n_u$ are identified by linearizing the continuous system dynamics of the considered benchmark, computing the discrete-time dynamics and eventually the transfer function in the z-domain.

\subsubsection{Bouc-Wen Oscillator}
The first dynamic system is the \textbf{Bouc-Wen} oscillator
\begin{subequations}
    \begin{align}
        m_L\ddot{y} &= u - k_Ly - c_L \dot{y} - z \\
        \dot{z} &= \alpha \dot{y} - \beta\left(\gamma |\dot{y}||z|^{\nu-1}z + \delta\dot{y}|z|^{\nu}\right)
    \end{align}
\end{subequations}
where the deplacement $y$ is the output, $u$ is an external force and $z$ is the hysteretic force.
With linearizing and discretizing the system dynamic the lags in~\eqref{eq:DynamicTarget} were identified as $n_y = 3$ and $n_u = 3$.
The regression task involves to find the change $\Delta y$ in the deplacement $y$ as it was defined in~\eqref{eq:DynamicTarget}.
For a comprehensive explanation for the system and the benchmark see~\citep{Schoukens2017}.

\subsubsection{Cascaded Tanks}
The second benchmark is a cascaded water two tank water system on top of each other, called \textbf{Tanks} in the following.
The nominal system dynamics are given by
\begin{subequations}
    \begin{align}
        \dot{x}_1 &= -k_1 \sqrt{x_1} + k_4 u + w_1 \\
        \dot{x}_2 &= k_2\sqrt{x_1} - k_3\sqrt{x_2} + w_2 \\
        y &= x_2 + e
    \end{align}
\end{subequations}
with the states $\bm{x}$, the control input $u$ and the output $y$.
The noise $w_1$ and $w_2$ models the input overflow, where water from the upper tank directly flows into the lower tank and/or directly into the reservoir below.
For the lag identification, $n_y = 2$ and $n_u = 2$ was obtained.
Further information can be found in~\citep{Schoukens2017}.

\subsubsection{Van der Pol Oscillator}
The third benchmark system is the well-known \textbf{Van der Pol} oscillator~\citep{Khalil2002}.
The differential equation is given by
\begin{equation}
    \label{eq:VanDerPol}
    \ddot{x} = \mu (1 - x^2)\dot{x} - x + w,
\end{equation}
where $\mu$ is the damping factor, $w = \Normal{0}{0.1}$ and $x$ is the output of the system~\citep{VanDerPol1926}.
The number of lags can directly be determined by the order of the system, which is $n_y = 2$ for $y = x$.
Moreover, since~\eqref{eq:VanDerPol} has no exogenous input,~\eqref{eq:DynamicTarget} simplifies to $y_k = y_{k-1} + f(y_{k-1}, y_{k-2})$.

The training and validation dataset were constructed by uniformly sampling 20 values from $[-2, 2]$ for the initial values for $x$ and $\dot{x}$.
Then from the initial values the system is simulated for $\qty{5}{\second}$ with a sampling time of $\mathrm{d}t = \qty{0.1}{\second}$.
This results in 1000 datapoints for training and validation.

\subsubsection{Building Energy System}
The fourth and last benchmark system is the thermal dynamics of one thermal zone, which is equivalent to a room in a \textbf{Building}.
It can be described by
\begin{subequations}
    \begin{align}
        C_z\dot{T}_z &= \frac{T_w-T_z}{R_{z}} + \frac{T_r-T_z}{R_{r}} + C_z w_1 \\
        C_w\dot{T}_w &= \frac{T_z-T_w}{R_{z}} + \frac{T_a-T_w}{R_{w}} + C_w w_2 \\
        C_r\dot{T}_r &= \frac{T_z-T_r}{R_{r}} + c_w \dot{m} (T_s-T_r) + C_r w_3
    \end{align}
\end{subequations}
with the zone, wall, radiator and ambient temperature $T_z$, $T_w$, $T_r$, and $T_a$.
The heat capacities and resistances are denoted by $C_*$, respectively $R_*$, where $*\in\left\{z,w,r\right\}$.
The control input is the hot water mass flow $\dot{m}$ with the specific heat capacity $c_w = \qty{4180}{\joule\per\kilogram\per\kelvin}$.
The hot water is supplied with $T_s = \qty{60}{\degreeCelsius}$.
Additionally, the noise $w_i = \Normal{0}{\num{5e-5}}$ is added to each equation to account for model uncertainties and unknown disturbances.
A detailed derivation for the system dynamics can be found in~\citep{Jemaa2018,MassaGray2016}.

The zone temperature $T_z$ is regulated by a PI controller to maintain the room temperature within a comfortable range of \qty{21}{\degreeCelsius} to \qty{24}{\degreeCelsius}.
During nights and weekends, this range is relaxed to \qty{17}{\degreeCelsius} to \qty{28}{\degreeCelsius}.
The dataset is computed by simulating the closed loop with the PI controller for one year with a sample time of $\mathrm{d}t = \qty{15}{\minute}$, where the training dataset is the first half of the year, and the validation dataset the second half, resulting in a training dataset size of 17518.

The number of lags in~\eqref{eq:DynamicTarget} was determined similar to the \textbf{Bouc-Wen} and \textbf{Tanks} examples for the inputs $u_1 = T_a$, and $u_2 = \dot{m}$, resulting in $n_{u_1} = 3$ and $n_{u_2} = 3$.
Since this example has multiple inputs,~\eqref{eq:DynamicTarget} has to be minimally adapted to account for $u_1$ and $u_2$.
Together with the zone temperature $T_z$, this results in 9 input features with the output $y = T_z$.
Moreover, since the scales of the input features vary in magnitude, the input features were scaled and an offset was removed.

\section{Offline Evaluation}
\label{sec:OfflineEvaluation}

The reduction criteria $f_{\mathrm{red}}(\mathcal{D}_*)$ in Algorithm~\ref{alg:online}, respectively Algorithm~\ref{alg:reduction} are not purely restricted for online applications.
They are also usable offline during training to select to most informative datapoints in a dataset.
Thus, before diving into the more involved online setting, the reduction and acceptance criteria are evaluated offline in different aspects.
This includes the computational complexity and computation times, followed by the reduction behavior.
In the end, the reduction criteria and their acceptance criteria are evaluated against the benchmarks introduced in the previous chapter.

\subsection{Computational Complexity}

Identifying the datapoint with the least relevance is normally not real-time capable, since a new GP has to be constructed for every datapoint~\citep{Kabzan2019}.
Nevertheless, the faster the relevance detection, the more datapoints can be considered in an online application.
The computational complexity of each criterion is analyzed using the big $\mathcal{O}$ notation given in Table~\ref{tab:Complexity}, which typically highlights only the dominant scaling term. 
However, this can obscure important differences between the algorithms, especially when sub-dominant terms are significant in practice. 
Therefore, in addition to the leading order term, lower order scaling terms are shown, while constant factors are omitted for clarity.

\begin{table}
    \caption{Comparison of the computational complexity of the different reduction criteria for a random dataset with $N=20$ and $N=100$ datapoints. The computation time is the median of 1000 function evaluations.}
    \label{tab:Complexity}
    \centering
    \begin{tabular}{l|c|c|c}
        Criterion & Complexity $\mathcal{O}$ & $N=20$ & $N=100$ \\
        \hline
        \emph{Prior Entropy}~\eqref{eq:prior_Entropy} & $\mathcal{O}\left(\frac{N^3}{6} + N^2 + N\right)$ & $\qty{0.054}{\milli\second}$ & $\qty{0.222}{\milli\second}$ \\
        \emph{Predictive Entropy}~\eqref{eq:pred_Entropy} & $\mathcal{O}\left(\frac{N^3}{6} + \frac{3}{2}N^2 + 2N\right)$ & $\qty{0.083}{\milli\second}$ & $\qty{0.259}{\milli\second}$ \\
        \emph{Mean Relevance}~\eqref{eq:MR} & $\mathcal{O}\left(\frac{N^3}{6} + 2N^2 + 3N\right)$ & $\qty{0.076}{\milli\second}$ &  $\qty{0.270}{\milli\second}$ \\
        \emph{Log Predictive Density}~\eqref{eq:LPD} & $\mathcal{O}\left(\frac{N^3}{6} + \frac{5}{2}N^2 + 3N\right)$ & $\qty{0.096}{\milli\second}$ & $\qty{0.300}{\milli\second}$ \\
        \emph{Marginal Log Likelihood}~\eqref{eq:MLL} & $\mathcal{O}\left(\frac{N^3}{6} + 2N^2 + 2N\right)$ & $\qty{0.068}{\milli\second}$ & $\qty{0.256}{\milli\second}$ \\
        \hline
    \end{tabular}
\end{table}

Table~\ref{tab:Complexity} summarizes the computational complexity and computation times for each criterion.
All criteria are primarily dominated by the Cholesky decomposition, which scales with $N^3/6$.
Additionally, the kernel matrix computation introduces a fixed $N^2$ scaling factor.
Computation times were measured in Python using NumPy and SciPy for $N=20$ and $N=100$ to give a general impression.
In both cases, \emph{Prior Entropy} is the fastest criterion.

In the following evaluations, only \emph{Prior Entropy}, \emph{Mean Relevance}, and \emph{Marginal Log Likelihood} are considered. This is because \emph{Prior Entropy} and \emph{Predictive Entropy} are equivalent in their reduction behavior, as are \emph{Marginal Log Likelihood} and \emph{Log Predictive Density}, see Section~\ref{sec:OnlineGaussianProcess} for details. 
Therefore, to avoid redundancy and remain consistent with the given literature, only one criterion from each pair is included in the subsequent analysis.

\subsection{Reduction Behavior}

\begin{figure}
    \centering
    \begin{tikzpicture}
\begin{groupplot}[
    group style={
        group size=4 by 1,
        horizontal sep=0.75cm,
        xlabels at=edge bottom,
        ylabels at=edge left,
        xticklabels at=edge bottom
    },
    width=4.4cm,
    height=4.5cm,
    xlabel={Number of datapoints},
    ylabel={SMSE},
    xlabel style={font=\footnotesize},
    ylabel style={font=\footnotesize},
    xticklabel style={font=\footnotesize},
    yticklabel style={font=\footnotesize},
    log ticks with fixed point,
    title style={font=\bfseries\small},
    xmin=1,
    ymode=log,
    xmax=100,
    xmajorgrids,
    xminorgrids,
    ymajorgrids,
    yminorgrids,
    minor x tick num=3,
    major grid style={gray!80},
    minor grid style={gray!40},
]
    \nextgroupplot[
        title={Himmelblau},
        ymin = 0.1,
    ]
        \addplot [blau, line width=1.2pt]
            table[x=x, y=PredictiveEntropy]{Himmelblau.dat};
        \addplot [orange, line width=1.2pt]
            table[x=x, y=MeanRelevance]{Himmelblau.dat};
        \addplot [gruen, line width=1.2pt]
            table[x=x, y=MLL]{Himmelblau.dat};
    \nextgroupplot[
        title={Rastrigin},
    ]
        \addplot [blau, line width=1.2pt]
            table[x=x, y=PredictiveEntropy]{Rastrigin.dat};
        \addplot [orange, line width=1.2pt]
            table[x=x, y=MeanRelevance]{Rastrigin.dat};
        \addplot [gruen, line width=1.2pt]
            table[x=x, y=MLL]{Rastrigin.dat};
    \nextgroupplot[
        title={Rosenbrock},
    ]
        \addplot [blau, line width=1.2pt]
            table[x=x, y=PredictiveEntropy]{Rosenbrock.dat};
        \addplot [orange, line width=1.2pt]
            table[x=x, y=MeanRelevance]{Rosenbrock.dat};
        \addplot [gruen, line width=1.2pt]
            table[x=x, y=MLL]{Rosenbrock.dat};
    \nextgroupplot[
        title={Six Hump Camel},
        legend to name=pgfplots:Iter,
        legend style={
            font=\footnotesize,
            cells={anchor=west},
        }
    ]
        \addplot [blau, line width=1.2pt]
            table[x=x, y=PredictiveEntropy]{Six_Hump_Camel.dat};
            \addlegendentry{Prior Entropy}
        \addplot [orange, line width=1.2pt]
            table[x=x, y=MeanRelevance]{Six_Hump_Camel.dat};
            \addlegendentry{Mean Relevance}
        \addplot [gruen, line width=1.2pt]
            table[x=x, y=MLL]{Six_Hump_Camel.dat};
            \addlegendentry{Marginal Log Likelihood}
\end{groupplot}%
\end{tikzpicture}
\pgfplotslegendfromname{pgfplots:Iter}%
%
    \vskip -1cm
    \caption{Comparison of the reduction behavior for the criteria. The SMSE is evaluated in every iteration during the reduction.}
    \label{fig:ReductionPlot}
\end{figure}
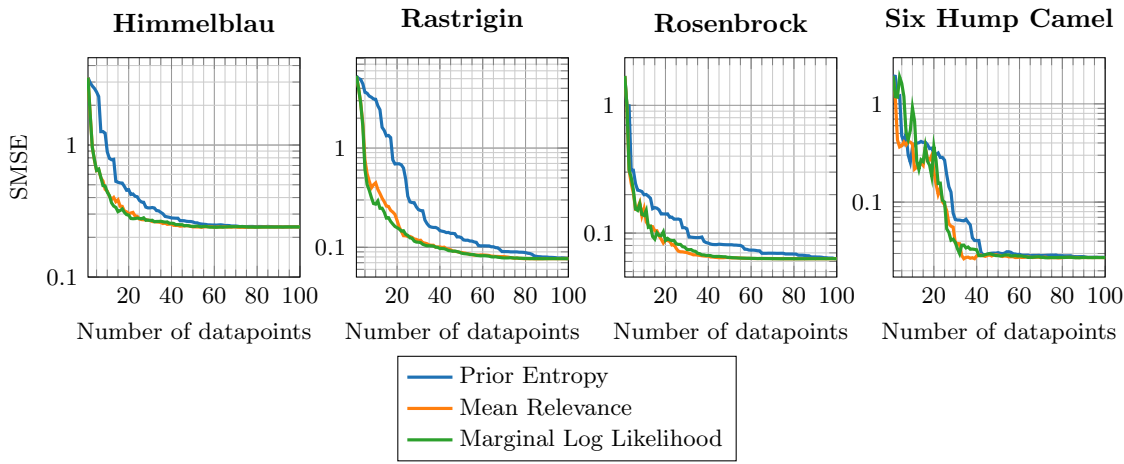

The first evaluation focuses on the reduction behavior of the criteria.
Only the benchmark functions are used, as the GP training datasets can be uniformly sampled from the function-specific grids.
This ensures comparability, since the training dataset includes the same region as the validation dataset.
The GPs are initially trained with $N=100$ uniformly sampled datapoints for each function.
After that, the number of datapoints is repeatedly reduced to one, evaluating the standardized mean squared error (SMSE) given by
\begin{equation}
    \mathrm{SMSE} = \frac{\frac{1}{N}\sum_{i=1}^{N}(y_i - \mu_i)^2}{\mathrm{Var}(\bm y)}
\end{equation}
for every iteration and successively reduced set size $N$.
A lower SMSE means a better prediction performance, where an $\mathrm{SMSE} = 1$ is the same as using the mean of $\bm{y}$ for $\mu_i$.
If $\mathrm{SMSE} > 1$, the prediction performance is worse than the mean.

The results are shown in Figure~\ref{fig:ReductionPlot}.
For \textbf{Rastrigin}, \textbf{Rosenbrock} and \textbf{Himmelblau} the results are similar. 
\emph{Prior Entropy} achieves a higher SMSE in every step. 
For \textbf{Rastrigin} and \textbf{Himmelblau} it also crosses $\mathrm{SMSE} = 1$ earlier than \emph{Mean Relevance} and \emph{Marginal Log Likelihood}.
On the other hand, \emph{Mean Relevance} and \emph{Marginal Log Likelihood} behave very similar in these cases.
For the \textbf{Six Hump Camel} function, the gap between \emph{Prior Entropy}, \emph{Marginal Log Likelihood} and \emph{Mean Relevance} is smaller. 
This difference is mainly due to a smaller variance in the target value in the \textbf{Six Hump Camel} function, leading to a lower prediction error for the mean in~\eqref{eq:pred_mu}.
Thus, the variance dominates in \emph{Marginal Log Likelihood}.

In summary, \emph{Prior Entropy} yields a higher SMSE.
Lower prediction errors are achieved by \emph{Marginal Log Likelihood} and \emph{Mean Relevance}.
Both criteria account for the actual function values in contrast to \emph{Prior Entropy}, removing datapoints whose relevance for the mean prediction is low.
\emph{Marginal Log Likelihood} yields a slightly higher SMSE than \emph{Mean Relevance} due to its additional consideration of the variance, which results in a lower standard deviation compared to \emph{Mean Relevance}.

Overall, \emph{Mean Relevance} and \emph{Marginal Log Likelihood} are preferable to \emph{Prior Entropy} when reducing the number of datapoints in a GP.
When focusing purely on datapoint reduction, \emph{Marginal Log Likelihood} does not exhibit a distinct advantage over \emph{Mean Relevance}.

\subsection{Acceptance Behavior}

The next evaluation concerns the acceptance behavior, for which all benchmark functions are used. 
Each benchmark is trained with 100 datapoints: for the grid-based benchmarks, 100 points are uniformly sampled, while for the benchmark datasets the first 100 training samples are selected. 
No insertion criterion is applied, so $\texttt{insert\_datapoint}(\mathcal{D}_*)$ always evaluates to true in Algorithm~\ref{alg:online}.

\begin{table*}
    \centering
    \caption{Resulting SMSE after updating the GP with 500 new datapoints. Initial shows the starting SMSE. The Accept column shows the SMSE when using the acceptance criterion for the respective reduction criterion. The last column shows the percentage of accepted datapoints. The lowest SMSE is marked in bold for normal operation and when using the acceptance criterion.}
    \label{tab:Acception}
    {\tiny
    \begin{tabular}{l|r|r||r|r|r|r|r|r|r|r|r}
        & Initial & N & \multicolumn{3}{c|}{\emph{Prior Entropy}} & \multicolumn{3}{c|}{\emph{Mean Relevance}} & \multicolumn{3}{c}{\emph{Marginal Log Likelihood}} \\
        \hline
        & SMSE & & SMSE & Accept & Perc. & SMSE & Accept & Perc. & SMSE & Accept & Perc. \\
        \hline
        Rastrigin & 0.0769 & 500 & 0.0513 & 0.0524 & 98.8\% & 0.0206 & 0.0208 & 89.4\% & \textbf{0.0166} & \textbf{0.0166} & 58.6\% \\ 
        Rosenbrock & 0.0631 & 500 & 0.0398 & 0.0414 & 95.2\% & \textbf{0.0207} & 0.0208 & 75.6\% & 0.0211 & \textbf{0.0203} & 50.4\% \\ 
        Himmelblau & 0.2396 & 500 & 0.2230 & 0.2221 & 94.0\% & 0.1109 & 0.1127 & 95.4\% & \textbf{0.0994} & \textbf{0.0990} & 44.8\% \\
        Six Hump Camel & 0.0273 & 500 & 0.0078 & 0.0082 & 98.8\% & 0.0044 & 0.0044 & 59.6\% & \textbf{0.0040} & \textbf{0.0040} & 93.4\% \\
        Boston & 0.4601 & 351 & 0.3559 & 0.3711 & 99.1\% & 0.2882 & 0.2928 & 93.4\% & \textbf{0.2681} & \textbf{0.2734} & 65.2\% \\
        Concrete & 0.9930 & 827 & 0.3911 & 0.3911 & 97.3\% & 0.4369 & 0.4429 & 96.7\% & \textbf{0.3041} & \textbf{0.2783} & 57.6\% \\ 
        Bouc-Wen & 0.0522 & 897 & 0.0287 & 0.0302 & 33.9\% & \textbf{0.0147} & \textbf{0.0144} & 86.2\% & 0.0287 & 0.0302 & 33.9\% \\
        Tanks & 0.7454 & 922 & 0.5164 & 0.5657 & 46.4\% & 0.4985 & 0.5150 & 95.2\% & \textbf{0.4621} & \textbf{0.4582} & 33.7\% \\
        Van der Pol & 0.0012 & 900 & 0.0007 & 0.0007 & 45.1\% & \textbf{0.0006} & \textbf{0.0006} & 36.4\% & 0.0007 & 0.0007 & 47.6\% \\
        Building & 0.9815 & 17418 & 0.7087 & 0.7125 & 6.7\% & 0.4706 & 0.4695 & 95.1\% & \textbf{0.4283} & \textbf{0.4162} & 3.6\% \\
        \hline
    \end{tabular}
    }
\end{table*}

The results are summarized in Table~\ref{tab:Acception}.
Overall, including new datapoints online results in a lower SMSE, implying a better prediction performance.
As in the previous section, \emph{Marginal Log Likelihood} and \emph{Mean Relevance} outperform \emph{Prior Entropy}.
For \emph{Prior Entropy}, the acceptance criterion does accept a main portion of the datapoints for the benchmark functions \textbf{Boston} and \textbf{Concrete}. 
For the system identification datasets, the acceptance percentage is lower while maintaining a similar SMSE.

\emph{Mean Relevance} achieves a slightly greater reduction in the number of included datapoints compared to \emph{Prior Entropy}, except for the \textbf{Six Hump Camel} function and \textbf{Van der Pol}, where 60\%, respectively 36\%, of the datapoints are accepted.
On the other hand, more datapoints are accepted for \textbf{Bouc-Wen} and \textbf{Building} which is compensated by a lower SMSE.

The best results are achieved using the acceptance criterion from \emph{Marginal Log Likelihood}, which results in an acceptance rate of about 50\% for \textbf{Rastrigin}, \textbf{Rosenbrock}, and \textbf{Himmelblau}, while maintaining similar or even better prediction performance. 
For the \textbf{Six Hump Camel} function, however, the acceptance rate is over 90\%.
Again, the predictive variance dominates the \emph{Marginal Log Likelihood}.

Overall, \emph{Marginal Log Likelihood} achieves the lowest SMSE in most cases and therefore the best overall prediction results.
Also, considering every datapoint in the training dataset results in an overall reduction for the SMSE, indicating the applicability of the reduction criteria as a data selection criteria during training.
Especially for the \textbf{Six Hump Camel} function and the \textbf{Concrete} and \textbf{Building} datasets, the SMSE was lowered by 60\% to 80\%.

\begin{figure*}
    \centering
    \includegraphics[width=0.8\textwidth]{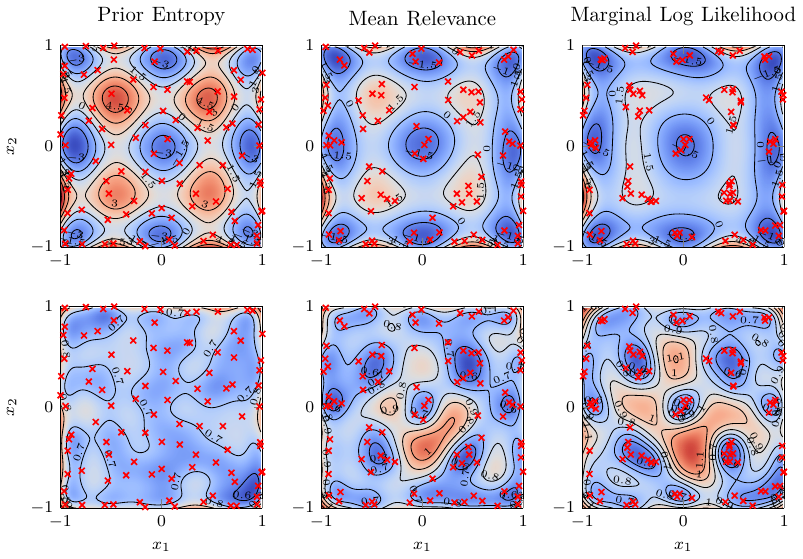}
    \caption{Prediction error and standard deviation of the \textbf{Rastrigin} function after seeing the training dataset. Red crosses mark the locations of the included datapoints.}
    \label{fig:Rastrigin}
\end{figure*}

Figure~\ref{fig:Rastrigin} examines the results for the \textbf{Rastrigin} function more closely by presenting the prediction error and standard deviation of the GP after processing the training dataset with the acceptance criterion.
The \emph{Prior Entropy} selects datapoints uniformly in the evaluation grid, with a strong focus on the boundaries.
This effectively reduces the predictive variance but neglects the prediction error.
This effect was already observed by~\citep{Krause2008,Ramakrishnan2005}.
In contrast, \emph{Mean Relevance} selects datapoints which reduce the prediction error, resulting in a neglection of the predictive variance.
Moreover, unlike \emph{Prior Entropy}, no focus on the grid boundaries is present and a clustering can be observed.
\emph{Marginal Log Likelihood} shows a similar behavior, but shows a more pronounced clustering as in \emph{Mean Relevance} and also grid boundary focus like in \emph{Prior Entropy}. 
This results in a further reduction of the predictive error which can be seen in Table~\ref{tab:Acception}.

\section{Online Evaluation}
\label{sec:OnlineEvaluation}

After the offline evaluation, the reduction and acceptance criteria are now evaluated in an online GP setting according to Algorithm~\ref{alg:online}.
We first start with an evaluation of all benchmark functions and datasets.
For \textbf{Van der Pol} and \textbf{Building}, an additional evaluation is conducted in which the error threshold is varied.

\subsection{General Benchmark Evaluations}

\setlength{\tabcolsep}{4pt}
\begin{table*}
    \centering
    \caption{Resulting SMSE after the application of the online GP in Algorithm~\ref{alg:online} with only the \textbf{variance threshold} $\bar{\sigma}^2$. Initial shows the starting SMSE. The Accept column shows the SMSE when using the acceptance criterion for the respective reduction criterion. The N column shows the number of datapoints which where considered during online training. The lowest SMSE is marked in bold for normal operation and when using the acceptance criterion.}
    \label{tab:OnlineVar}
    {\tiny
    \begin{tabular}{l|r|r||r|r|r|r|r|r|r|r|r|r|r|r}
        & Initial & \multicolumn{1}{c||}{$\bar{\sigma}^2$} & \multicolumn{4}{c|}{\emph{Prior Entropy}} & \multicolumn{4}{c|}{\emph{Mean Relevance}} & \multicolumn{4}{c}{\emph{Marginal Log Likelihood}} \\
        \hline
        & & & \multicolumn{2}{c|}{Normal} & \multicolumn{2}{c|}{Accept} & \multicolumn{2}{c|}{Normal} & \multicolumn{2}{c|}{Accept} & \multicolumn{2}{c|}{Normal} & \multicolumn{2}{c}{Accept} \\
        \hline
        & SMSE & & SMSE & N & SMSE & N & SMSE & N & SMSE & N & SMSE & N & SMSE & N \\
        \hline
        Rastrigin & 0.0769 & 0.5 & 0.0493 & 153 & 0.0493 & 153 & 0.0289 & 210 & 0.0295 & 190 & \textbf{0.0248} & 238 & \textbf{0.0256} & 151 \\ 
        Rosenbrock & 0.0631 & 0.5 & 0.0417 & 57 & 0.0417 & 57 & 0.0309 & 149 & \textbf{0.0301} & 125 & \textbf{0.0307} & 135 & 0.0305 & 81 \\ 
        Himmelblau & 0.2396 & 0.5 & 0.2001 & 51 & 0.2001 & 51 & 0.1366 & 85 & 0.1371 & 72 & \textbf{0.1315} & 79 & \textbf{0.1320} & 66 \\
        Six Hump Camel & 0.0273 & 0.01 & 0.0074 & 40 & 0.0074 & 40 & 0.0067 & 45 & 0.0065 & 40 & \textbf{0.0063} & 41 & \textbf{0.0063} & 41 \\
        Boston & 0.4601 & 1.5 & 0.3700 & 168 & 0.3700 & 168 & \textbf{0.2733} & 157 & \textbf{0.2733} & 152 & 0.2938 & 153 & 0.2938 & 127 \\
        Concrete & 0.9930 & 2.0 & 0.4460 & 260 & 0.4460 & 260 & 0.3465 & 261 & \textbf{0.3584} & 255 & \textbf{0.3420} & 282 & 0.3854 & 219 \\
        BoucWen & 0.0522 & 0.001 & 0.0311 & 70 & 0.0311 & 70 & \textbf{0.0235} & 77 & \textbf{0.0240} & 75 & 0.0311 & 70 & 0.0311 & 70 \\ 
        Tanks & 0.7454 & 0.003 & 0.5263 & 20 & 0.5263 & 20 & \textbf{0.5197} & 21 & \textbf{0.5163} & 19 & 0.5233 & 20 & 0.5233 & 20 \\
        Van der Pol & 0.0019 & 0.00075 & 0.0015 & 39 & 0.0015 & 39 & 0.0015 & 50 & 0.0015 & 21 & \textbf{0.0015} & 39 & \textbf{0.0015} & 39 \\
        Building & 0.9815 & 0.0015 & 0.7592 & 16 & 0.7592 & 16 & 0.7491 & 17 & 0.7491 & 17 & \textbf{0.7289} & 17 & \textbf{0.7289} & 17 \\
        \hline
    \end{tabular}
    }
\end{table*}

\begin{table*}
    \centering
    \caption{Resulting SMSE after the application of the online GP in Algorithm~\ref{alg:online} with only the \textbf{error threshold} $\bar{e}$. Initial shows the starting SMSE. The Accept column shows the SMSE when using the acceptance criterion for the respective reduction criterion. The N column shows the number of datapoints which where considered during online training. The lowest SMSE is marked in bold for normal operation and when using the acceptance criterion.}
    \label{tab:OnlineErr}
    {\tiny
    \begin{tabular}{l|r|r||r|r|r|r|r|r|r|r|r|r|r|r}
        & Initial & \multicolumn{1}{c||}{$\bar{e}$} & \multicolumn{4}{c|}{\emph{Prior Entropy}} & \multicolumn{4}{c|}{\emph{Mean Relevance}} & \multicolumn{4}{c}{\emph{Marginal Log Likelihood}} \\
        \hline
        & & & \multicolumn{2}{c|}{Normal} & \multicolumn{2}{c|}{Accept} & \multicolumn{2}{c|}{Normal} & \multicolumn{2}{c|}{Accept} & \multicolumn{2}{c|}{Normal} & \multicolumn{2}{c}{Accept} \\
        \hline
        & SMSE & & SMSE & N & SMSE & N & SMSE & N & SMSE & N & SMSE & N & SMSE & N \\
        \hline
        Rastrigin & 0.0769 & 2.5 & 0.0387 & 119 & 0.0394 & 116 & 0.0232 & 68 & 0.0232 & 68 & \textbf{0.0199} & 67 & \textbf{0.0199} & 67 \\ 
        Rosenbrock & 0.0631 & 25.0 & 0.0315 & 39 & 0.0315 & 37 & 0.0253 & 26 & 0.0253 & 26 & \textbf{0.0238} & 25 & \textbf{0.0238} & 25 \\ 
        Himmelblau & 0.2396 & 25.0 & 0.1709 & 147 & 0.1738 & 141 & 0.1167 & 101 & 0.1167 & 101 & \textbf{0.1017} & 88 & \textbf{0.1017} & 88 \\
        Six Hump Camel & 0.0273 & 0.5 & 0.0104 & 4 & 0.0104 & 4 & 0.0104 & 4 & 0.0104 & 4 & \textbf{0.0103} & 4 & \textbf{0.0103} & 4 \\ 
        Boston & 0.4601 & 2.5 & 0.3705 & 155 & 0.3705 & 155 & 0.2961 & 163 & 0.2961 & 163 & \textbf{0.2693} & 155 & \textbf{0.2693} & 155 \\ 
        Concrete & 0.9930 & 5.0 & 0.4163 & 459 & 0.3815 & 451 & 0.3900 & 469 & 0.3900 & 469 & \textbf{0.2848} & 454 & \textbf{0.2561} & 412 \\
        BoucWen & 0.0522 & 5e-05 & 0.0224 & 152 & 0.0236 & 113 & \textbf{0.0143} & 112 & \textbf{0.0143} & 112 & 0.0224 & 152 & 0.0236 & 113 \\ 
        Tanks & 0.7454 & 0.05 & \textbf{0.4634} & 247 & \textbf{0.4735} & 192 & 0.5107 & 264 & 0.5107 & 264 & 0.4642 & 255 & 0.4752 & 224 \\
        Van der Pol & 0.0019 & 0.005 & 0.0006 & 230 & 0.0006 & 182 & 0.0006 & 253 & 0.0006 & 213 & \textbf{0.0006} & 212 & \textbf{0.0006} & 178 \\ 
        Building & 0.9815 & 0.2 & 0.4666 & 31 & 0.4666 & 31 & \textbf{0.4387} & 34 & \textbf{0.4387} & 34 & 0.4439 & 31 & 0.4439 & 31 \\
        \hline
    \end{tabular}
    }
\end{table*}

Table~\ref{tab:OnlineVar} shows the results when using only the variance insertion criterion.
As expected, the SMSE is higher than in Table~\ref{tab:Acception}, since the insertion criterion filters out unsuitable datapoints.
Differently, \emph{Mean Relevance} performs better for half of the test cases than \emph{Marginal Log Likelihood}, but with revising more datapoints.

The results for using the error threshold $\bar{e}$ are shown in Table~\ref{tab:OnlineErr}.
Compared to Table~\ref{tab:OnlineVar}, the SMSE is generally lower.
For half the benchmarks, this is accompanied by a lower number of considered datapoints.
When looking at the datasets \textbf{Concrete}, \textbf{Boston}, \textbf{Bouc-Wen} and \textbf{Tanks}, the higher number of datapoints shows that the corresponding regression task is more involved and maybe additional input features or a higher budget $N_{max}$ is needed.

\subsection{Van der Pol oscillator}

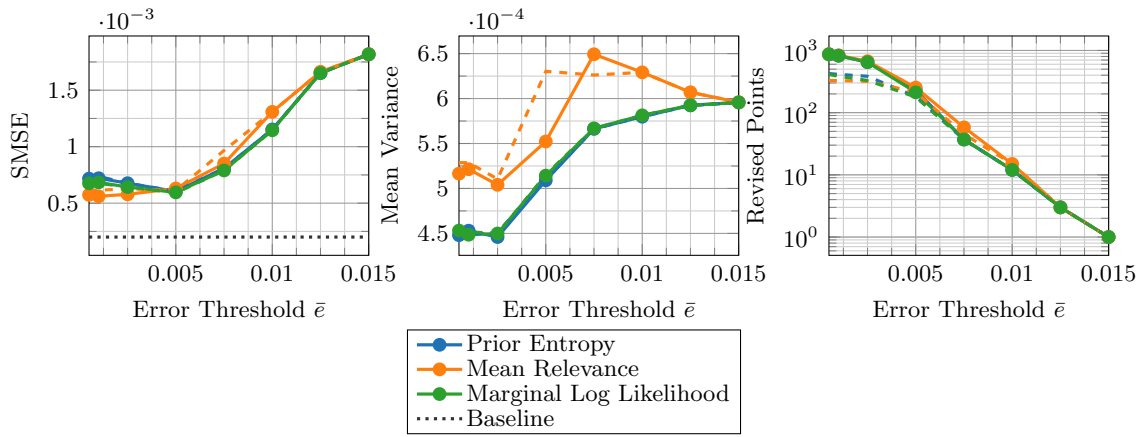
\begin{figure}
    \centering
    \begin{tikzpicture}
\begin{groupplot}[
    group style={
        group size=3 by 1,
        vertical sep=0.25cm,
        horizontal sep=1.2cm,
        xlabels at=edge bottom,
        ylabels at=edge left,
        xticklabels at=edge bottom
    },
    width=5.3cm,
    height=4.5cm,
    xlabel={Error Threshold $\bar{e}$},
    xlabel style={font=\footnotesize},
    ylabel style={font=\footnotesize},
    scaled x ticks=false,
    scaled y ticks=false,
    xticklabel style={
        font=\footnotesize,
        /pgf/number format/fixed,
        /pgf/number format/precision=4,
    },
    xmin=0.0005,
    xmax=0.015,
    xmajorgrids,
    xminorgrids,
    ymajorgrids,
    yminorgrids,
    minor x tick num=3,
    major grid style={gray!80},
    minor grid style={gray!40},
]
    \nextgroupplot[
        ylabel={SMSE},
        legend to name=pgfplots:VanDerPol,
        legend style={
            font=\footnotesize,
            inner ysep=1pt,
            nodes={inner sep=1pt,text depth=0.1em},
            cells={anchor=west},
        },
        yticklabel style={
            font=\footnotesize,
        },
        scaled y ticks=base 10:3,
        minor y tick num=1,
    ]
        \addplot [blau, line width=1.2pt, mark=*]
            table[x=x, y=PredictiveEntropy]{VanDerPol_Error_SMSE.dat};
            \addlegendentry{Prior Entropy};
        \addplot [orange, line width=1.2pt, mark=*]
            table[x=x, y=MeanRelevance]{VanDerPol_Error_SMSE.dat};
            \addlegendentry{Mean Relevance};
        \addplot [gruen, line width=1.2pt, mark=*]
            table[x=x, y=MLL]{VanDerPol_Error_SMSE.dat};
            \addlegendentry{Marginal Log Likelihood};
        \addplot [black!80, dotted, line width=1.2pt]
            coordinates {(0.0005, 0.0002) (0.015, 0.0002)};
            \addlegendentry{Baseline};
        \addplot [blau, dashed, line width=1.2pt]
            table[x=x, y=PredictiveEntropy]{VanDerPol_Reject_Error_SMSE.dat};
        \addplot [orange, dashed, line width=1.2pt]
            table[x=x, y=MeanRelevance]{VanDerPol_Reject_Error_SMSE.dat};
        \addplot [gruen, dashed, line width=1.2pt]
            table[x=x, y=MLL]{VanDerPol_Reject_Error_SMSE.dat};
    \nextgroupplot[
        ylabel={Mean Variance},
        yticklabel style={
            font=\footnotesize,
        },
        scaled y ticks=base 10:4,
        ytick={0.00045, 0.0005, 0.00055, 0.0006, 0.00065},
        minor y tick num=1,
    ]
        \addplot [blau, line width=1.2pt, mark=*]
            table[x=x, y=PredictiveEntropy]{VanDerPol_Error_Var.dat};
        \addplot [orange, line width=1.2pt, mark=*]
            table[x=x, y=MeanRelevance]{VanDerPol_Error_Var.dat};
        \addplot [gruen, line width=1.2pt, mark=*]
            table[x=x, y=MLL]{VanDerPol_Error_Var.dat};
        \addplot [blau, dashed, line width=1.2pt]
            table[x=x, y=PredictiveEntropy]{VanDerPol_Reject_Error_Var.dat};
        \addplot [orange, dashed, line width=1.2pt]
            table[x=x, y=MeanRelevance]{VanDerPol_Reject_Error_Var.dat};
        \addplot [gruen, dashed, line width=1.2pt]
            table[x=x, y=MLL]{VanDerPol_Reject_Error_Var.dat};
    \nextgroupplot[
        ymode=log,
        ylabel={Revised Points},
        yticklabel style={
            font=\footnotesize,
        },
    ]
        \addplot [blau, line width=1.2pt, mark=*]
            table[x=x, y=PredictiveEntropy]{VanDerPol_Error_N.dat};
        \addplot [orange, line width=1.2pt, mark=*]
            table[x=x, y=MeanRelevance]{VanDerPol_Error_N.dat};
        \addplot [gruen, line width=1.2pt, mark=*]
            table[x=x, y=MLL]{VanDerPol_Error_N.dat};
        \addplot [blau, dashed, line width=1.2pt]
            table[x=x, y=PredictiveEntropy]{VanDerPol_Reject_Error_N.dat};
        \addplot [orange, dashed, line width=1.2pt]
            table[x=x, y=MeanRelevance]{VanDerPol_Reject_Error_N.dat};
        \addplot [gruen, dashed, line width=1.2pt]
            table[x=x, y=MLL]{VanDerPol_Reject_Error_N.dat};

\end{groupplot}%
\end{tikzpicture}
\pgfplotslegendfromname{pgfplots:VanDerPol}%
%
    \vskip -1cm
    \caption{SMSE, mean predictive variance and revised datapoints for the online GP for the Van der Pol oscillator. Here only the error threshold as an insertion criterion was used. The dashed colored lines show the results when using the acceptance criteria.}
    \label{fig:VanDerPolError}
\end{figure}

The influence of the error threshold is investigated in more detail for the \textbf{Van der Pol} oscillator and the \textbf{Building} dataset (see Section~\ref{sec:Building}).
For a baseline, a full GP was trained with the whole training dataset, consisting out of 1000 datapoints, resulting in a SMSE of 0.0002.

The results for using only the error threshold is shown in Figure~\ref{fig:VanDerPolError}.
All reduction criteria show a similar trend in their SMSE and in the number of revised datapoints. 
For $\bar{e} \geq 0.005$ \emph{Mean Relevance} has a higher SMSE and number of revised datapoints than the other criteria.
For lower values, the situation for the SMSE switches probably due to the more datapoints \emph{Mean Relevance} can choose from.
The similarity between \emph{Marginal Log Likelihood} and \emph{Prior Entropy} shows a lower impact of the prediction error, which can also be seen when looking at the number of revised points when using an acceptance criterion. 
Here they also behave almost equal.
For \emph{Mean Relevance}, the acceptance criterion results in a slightly worse result, while also reducing the number of revised datapoints.

For $\bar{e} \leq 0.005$, \emph{Mean Relevance} achieves a lower SMSE than \emph{Prior Entropy} and \emph{Marginal Log Likelihood}.
Moreover, the SMSE for \emph{Prior Entropy} and \emph{Marginal Log Likelihood} rises compared to $\bar{e} = 0.005$, showing a worse result.
Nevertheless, both reduction criteria reduce the mean predictive variance, showing that the uncertainty dominates over the mean prediction error.
This can also be seen from the $\mathrm{SMSE} < 0.0015$, which is already very low.

Overall, the online GP achieves a comparable result to the full GP, considering the 10 times lower number of training datapoints, especially with the general low SMSE.

\subsection{Building}
\label{sec:Building}

For the \textbf{Building} dataset, also only the error threshold $\bar{e}$ as an insertion criterion was used. 
As a baseline comparison a sparse GP with the VFE approximation was trained with the entire training dataset and 100 inducing points, since the training set has 17518 instances.

\begin{figure}
    \centering
    \begin{tikzpicture}
\begin{groupplot}[
    group style={
        group size=3 by 1,
        vertical sep=0.25cm,
        horizontal sep=1.4cm,
        xlabels at=edge bottom,
        ylabels at=edge left,
        xticklabels at=edge bottom
    },
    width=5.1cm,
    height=4.5cm,
    xlabel={Error Threshold $\bar{e}$},
    xlabel style={font=\footnotesize},
    ylabel style={font=\footnotesize},
    scaled x ticks=false,
    scaled y ticks=false,
    xticklabel style={
        font=\footnotesize,
        /pgf/number format/fixed,
        /pgf/number format/precision=4,
    },
    xmin=0.05,
    xmax=0.25,
    xmajorgrids,
    xminorgrids,
    ymajorgrids,
    yminorgrids,
    minor x tick num=3,
    major grid style={gray!80},
    minor grid style={gray!40},
]
    \nextgroupplot[
        ylabel={SMSE},
        legend to name=pgfplots:Building,
        legend style={
            font=\footnotesize,
            inner ysep=1pt,
            nodes={inner sep=1pt,text depth=0.1em},
            cells={anchor=west},
        },
        yticklabel style={
            font=\footnotesize,
            /pgf/number format/fixed,
            /pgf/number format/precision=7,
        },
        minor y tick num=3,
    ]
        \addplot [blau, line width=1.2pt, mark=*]
            table[x=x, y=PredictiveEntropy]{Building_Error_SMSE.dat};
            \addlegendentry{Prior Entropy};
        \addplot [orange, line width=1.2pt, mark=*]
            table[x=x, y=MeanRelevance]{Building_Error_SMSE.dat};
            \addlegendentry{Mean Relevance};
        \addplot [gruen, line width=1.2pt, mark=*]
            table[x=x, y=MLL]{Building_Error_SMSE.dat};
            \addlegendentry{Marginal Log Likelihood};
        \addplot [black!80, dotted, line width=1.2pt]
            coordinates {(0.05, 0.3397) (0.25, 0.3397)};
            \addlegendentry{Baseline};
        \addplot [blau, dashed, line width=1.2pt]
            table[x=x, y=PredictiveEntropy]{Building_Reject_Error_SMSE.dat};
        \addplot [orange, dashed, line width=1.2pt]
            table[x=x, y=MeanRelevance]{Building_Reject_Error_SMSE.dat};
        \addplot [gruen, dashed, line width=1.2pt]
            table[x=x, y=MLL]{Building_Reject_Error_SMSE.dat};
    \nextgroupplot[
        ylabel={Mean Variance},
        yticklabel style={
            font=\footnotesize,
        },
        scaled y ticks=base 10:3,
        ytick={0.00025, 0.0005, 0.00075, 0.001, 0.00125, 0.0015},
        minor y tick num=1,
    ]
        \addplot [blau, line width=1.2pt, mark=*]
            table[x=x, y=PredictiveEntropy]{Building_Error_Var.dat};
        \addplot [orange, line width=1.2pt, mark=*]
            table[x=x, y=MeanRelevance]{Building_Error_Var.dat};
        \addplot [gruen, line width=1.2pt, mark=*]
            table[x=x, y=MLL]{Building_Error_Var.dat};
        \addplot [blau, dashed, line width=1.2pt]
            table[x=x, y=PredictiveEntropy]{Building_Reject_Error_Var.dat};
        \addplot [orange, dashed, line width=1.2pt]
            table[x=x, y=MeanRelevance]{Building_Reject_Error_Var.dat};
        \addplot [gruen, dashed, line width=1.2pt]
            table[x=x, y=MLL]{Building_Reject_Error_Var.dat};
    \nextgroupplot[
        ymode=log,
        ylabel={Revised Points},
        yticklabel style={
            font=\footnotesize,
        },
    ]
        \addplot [blau, line width=1.2pt, mark=*]
            table[x=x, y=PredictiveEntropy]{Building_Error_N.dat};
        \addplot [orange, line width=1.2pt, mark=*]
            table[x=x, y=MeanRelevance]{Building_Error_N.dat};
        \addplot [gruen, line width=1.2pt, mark=*]
            table[x=x, y=MLL]{Building_Error_N.dat};
        \addplot [blau, dashed, line width=1.2pt]
            table[x=x, y=PredictiveEntropy]{Building_Reject_Error_N.dat};
        \addplot [orange, dashed, line width=1.2pt]
            table[x=x, y=MeanRelevance]{Building_Reject_Error_N.dat};
        \addplot [gruen, dashed, line width=1.2pt]
            table[x=x, y=MLL]{Building_Reject_Error_N.dat};

\end{groupplot}%
\end{tikzpicture}
\pgfplotslegendfromname{pgfplots:Building}%
    \vskip -1cm
    \caption{SMSE, mean predictive variance and revised datapoints for the online GP for the Building dataset. Here only the error threshold as an insertion criterion was used. The dashed colored lines show the results when using the acceptance criteria.}
    \label{fig:Building}
\end{figure}
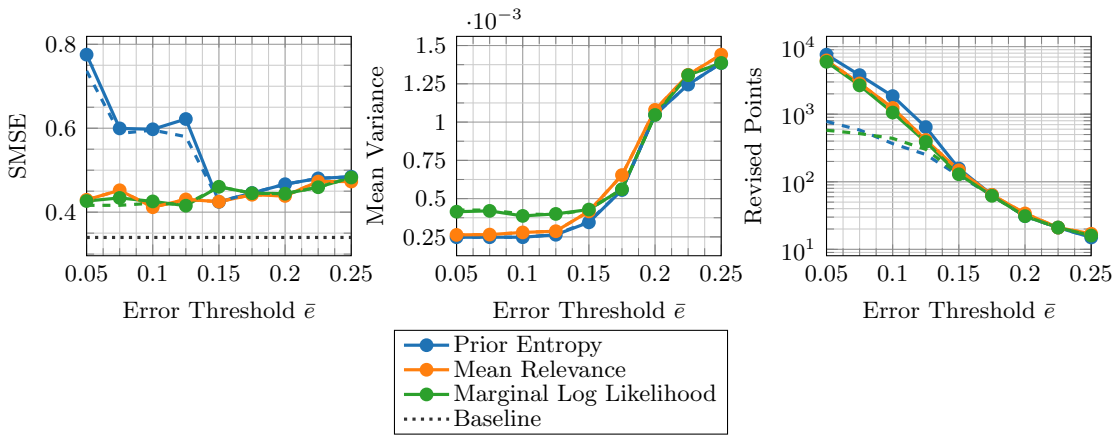

The results are shown in Figure~\ref{fig:Building} with the error threshold $\bar{e}$.
As expected, the number of revised datapoints is higher in general, if the error threshold is lower.
For the SMSE, no such monotone trend can be observed. 
For $\bar{e} \geq \qty{0.15}{\kelvin}$ all reduction criteria behave similarly with no substantial difference.
With lower error thresholds, the SMSE does not decrease further.
Moreover, \emph{Mean Relevance} and \emph{Marginal Log Likelihood} show an oscillating behavior in the SMSE.
Using an acceptance criterion smoothes this behavior for \emph{Marginal Log Likelihood}, while reducing the number of revised datapoints.
For \emph{Mean Relevance} no such behavior can be observed.
When looking at \emph{Prior Entropy}, the SMSE rises tremendously.
A similar behavior, but not that pronounced, was seen for the Van der Pol oscillator.
Again, \emph{Prior Entropy} does not account for the prediction error but only the predictive variance, which is correctly minimized.

Another interesting behavior can be seen in the number of revised points when using an acceptance criterion.
For $\bar{e} \leq \qty{0.125}{\kelvin}$ the graphs for \emph{Prior Entropy} and \emph{Marginal Log Likelihood} start to flatten out, showing that the acceptance criterion filters out points which contain redundant information.
Combined with no further decrease of the SMSE, this saturation shows the maximum achievable SMSE with the current budget $N_{max}$ for the given hyperparameters.

The online trained GPs performs quite well in comparison to the baseline sparse GP. 
They are for the best results only 0.1 higher than the baseline GP.
Moreover, the GPs only have to store $n^2$ datapoints where $n = 100$, whereas the sparse GP memory scaling is $m^2n$ with $m = 100$ and $n = 17518$. 
The lower prediction error of the sparse GP is no competition for the online GP considering the higher memory demand.

\section{Discussion}
\label{sec:Discussion}

As seen from the offline and online evaluations, using the data selection outlined in Algorithm~\ref{alg:online} and Algorithm~\ref{alg:reduction} results in a lower prediction error compared to the initial training.
This is especially true for the online setting, if the variance or error thresholds $\bar{\sigma}^2$ and $\bar{e}$ are low enough. 
For the offline data selection this leverages an efficient way to include the most informative datapoints while maintaining the budget $N_{max}$.
When looking at the baseline GPs for \textbf{Van der Pol} and \textbf{Building}, online learning or data selection results in a higher SMSE, but with greater data efficiency.
This greater efficiency also reduces the computation time for predictive mean and variance given in~\eqref{eq:pred_mu} and~\eqref{eq:pred_Sigma}.

When the acceptance criteria for \emph{Marginal Log Likelihood} and \emph{Prior Entropy} start to filter out datapoints, the chosen threshold for insertion cannot be satisfied with the current budget $N_{max}$.
On the other hand, the predictive performance does not worsen tremendously, showing that using the acceptance criteria for \emph{Marginal Log Likelihood} and \emph{Prior Entropy} has a positive impact and reduces the computational load.
However, for the offline data selection, the acceptance criterion for \emph{Marginal Log Likelihood} outperforms \emph{Prior Entropy}.

Using \emph{Prior Entropy} results in a higher SMSE compared to the other reduction criteria for nearly every benchmark.
Moreover, when too many datapoints are considered, the SMSE can rise again.
Thus, \emph{Prior Entropy} should only be used when the predictive error is of low concern.

The \emph{Mean Relevance} and \emph{Marginal Log Likelihood} criteria do not exhibit this behavior and are therefore preferable.
They show a more robust behavior for the SMSE since they are directly considering the predictive error.
Unlike in the offline evaluation, \emph{Mean Relevance} does show a comparable performance and depending on the use case, better prediction performance as \emph{Marginal Log Likelihood}.
This is mainly due to \emph{Marginal Log Likelihood} considering the uncertainty of the predictions.
If the uncertainty is weighted higher than the prediction error, \emph{Marginal Log Likelihood} results in a higher SMSE.

Overall, if only the prediction error is of concern, \emph{Mean Relevance} is the best choice.
On the other hand, if the uncertainty shall be included into the datapoint selection, \emph{Marginal Log Likelihood} should be chosen.

This can be summarized to the following guidelines:
\begin{itemize}
    \item For offline data selection, \emph{Marginal Log Likelihood} is the best choice.
    \item \emph{Mean Relevance} is the preferred criterion in an online setting if only the prediction error is of concern.
    \item \emph{Marginal Log Likelihood} / \emph{Log Predictive Density} balances the predictive variance with the predictive error. These criteria should be chosen, if the uncertainty is of concern.
    \item \emph{Prior Entropy} / \emph{Predictive Entropy} are only recommended when uniform sampling of the observation space is required and the prediction error is not a concern.
    \item The acceptance criterion for \emph{Marginal Log Likelihood} effectively filters out datapoints that provide no additional information content. 
    \item Overall, an acceptance criterion is recommended to reduce the computational demand.
\end{itemize}
\section{Conclusion} 
\label{sec:conclusion}

This work compared data reduction criteria for Gaussian Processes.
These criteria are mainly used in online GP algorithms with a fixed datapoint budget, but can also be used for data selection during offline training.
Such online algorithms typically consist out of an insertion and reduction criterion.
Additionally, acceptance criteria were proposed in this paper to filter out datapoints proposed by the insertion criterion, which do not add more information for the GP.

The criteria were compared based on their computational complexity, reduction behavior, and performance within the presented online GP algorithm.
Following the guidelines summarized in the previous chapter, \emph{Mean Relevance} is the best reduction criterion in an online setting.
When only offline data selection is conducted, \emph{Marginal Log Likelihood} achieved the best results.
The \emph{Prior Entropy} criterion can only be recommended, if a uniform sampling of the observation space is required.


With the presented acceptance criteria for \emph{Prior Entropy} and \emph{Marginal Log Likelihood}, unsuitable datapoints are consistently filtered out, thus leading to no evaluation of the reduction criterion.
If a great portion of datapoints are filtered out, this indicates that the insertion threshold can't be met with the current budget $N_{max}$.
In future research, the acceptance criteria can also be compared to each other and not only for their respective reduction criterion.

The reduction criteria can also be suitable for sparse GPs, as shown in~\citep{Kabzan2019} which used essentially \emph{Predictive Entropy}.
A throughout investigation and comparison for sparse GPs is topic of future research, since the handling of the inducing datapoints is more involved.
Also due to the approximation in sparse GPs, the equivalence between \emph{Prior Entropy} and \emph{Predictive Entropy} or \emph{Marginal Log Likelihood} and \emph{Log Predictive Density} does not hold anymore.

Another topic concerns parametric mean functions which can be jointly learned with the hyperparameters of the GP.
Replacing datapoints in the training set will lead to different results for the parameters of the mean function.
How this replacement can be considered, is an open topic.
Furthermore, a comparison between the recursive GPs from~\citep{Huber2014} can provide more insight for this problem and also highlights the performance between the recursive approach compared to the insertion/reduction approach used in this work.

\acks{Funded by the German Federal Ministry for Economic Affairs and Energy under grant number 03EN1066B.}

\vskip 0.2in
\bibliography{ref}

\end{document}